\documentclass{article}
\usepackage{spconf,amsmath,graphicx}
\usepackage{indentfirst}
\usepackage{multirow}
\usepackage{booktabs}
\usepackage[table]{xcolor}
\usepackage{subcaption}
\usepackage[skip=0pt]{caption}
\usepackage{amsfonts} 
\usepackage{bbding}

\title{MFPNet: Multi-scale Feature Propagation Network for 
Lightweight Semantic Segmentation}
\name{Guoan Xu$^{\star}$ \qquad Wenjing Jia$^{\star}$ \qquad Tao Wu$^{\dagger}$ \qquad Ligeng Chen$^{\dagger}$}
  
  \address{$^{\star}$ University of Technology Sydney\\
      $^{\dagger}$ Nanjing University }
\begin{document}
%
\maketitle
\begin{abstract}
In contrast to the abundant research focusing on large-scale models, 
the progress in 
lightweight semantic segmentation appears to be advancing at a 
comparatively slower pace. However, existing compact methods often 
suffer from limited 
feature representation capability due to the shallowness of their networks.
In this paper, we propose a novel lightweight segmentation architecture, called  Multi-scale Feature Propagation Network (MFPNet), to address the dilemma. 
Specifically, we design a robust 
Encoder-Decoder structure featuring 
symmetrical residual blocks that consist of 
flexible bottleneck residual modules (BRMs) to explore deep and rich muti-scale 
semantic context. Furthermore, taking benefit from their capacity to model latent long-range contextual relationships, we leverage Graph Convolutional Networks (GCNs) to facilitate multi-scale feature propagation between the BRM blocks. 
When evaluated on benchmark datasets, our proposed approach shows superior segmentation results.

\end{abstract}

\vspace{-0.5em}

\section{Introduction}
\label{sec:intro}

Semantic segmentation is an essential task in computer vision, which involves the classification of individual pixels 
into distinct semantic categories. 
Existing large-scale models, such as~\cite{chen2018encoder,zheng2021rethinking,lai2023denoising}, 
have undeniably 
achieved remarkable performance. 
However, they are far from meeting the requirements of resource-limited devices in terms of inference speed and computational complexity. 
As a result, researchers have increasingly turned their attention to compact methods as a means to alleviate 
the computational burden associated with these models. 
For example, models like ERFNet~\cite{romera2017erfnet} and DABNet~\cite{li2019dabnet} have designed non-bottleneck factorized convolution modules to reduce the convolutional dimension. However, these approaches sacrifice 
certain spatial details, thereby leading to suboptimal accuracy. 
Some more recent approaches such as DFANet~\cite{li2019dfanet}, ICNet~\cite{zhao2018icnet}, and FBSNet~\cite{gao2022fbsnet} have aimed 
to address this accuracy issue by employing a multi-branch architecture. However, using multiple branches can potentially introduce additional  
delays in the backpropagation process. 
In addition, methods like MSFNet~\cite{si2019real} and DSNet~\cite{chen2020dsnet} have adopted U-shaped architectures to utilize 
the Encoder's features to guide the resolution recovery process of the Decoder.  Nevertheless, the conventional convolutions used in the connection process can only aggregate spatially adjacent pixels, resulting in imperfect guidance.

\begin{figure}[t]
	\centerline{\includegraphics[width = 8.2cm]{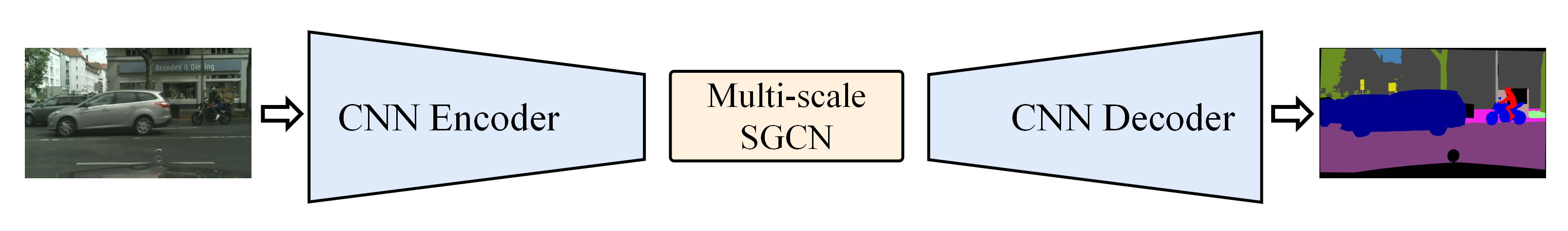}}
	\caption{The overall architecture of the proposed  Multi-scale Feature Propagation Network (MFPNet) method.}
	\label{Figure 1}
\vspace{-1.0em}
\end{figure}

To address the aforementioned challenges, in this work we propose a multi-scale feature propagation approach for lightweight semantic segmentation, abbreviated as MFPNet (see Fig.~\ref{Figure 1}), which considers both model capacity and inference speed. 
Our approach adopts 
a symmetrical codec structure enhanced with multi-scale Simple Graph Convolutional Networks (GCNs),  which enables effective propagation of latent object feature information. 
Compared to the limited local perception of the conventional 
convolution, our multi-scale feature propagation approach can aggregate information from spatially non-adjacent, long-range pixels belonging 
to the same class. This empowers 
a more comprehensive gathering of pixel-level details and contextual information, leading to superior segmentation results. 

\begin{figure*}[t]
        \centerline{\includegraphics[width=0.98\textwidth,height = 0.38\textwidth]{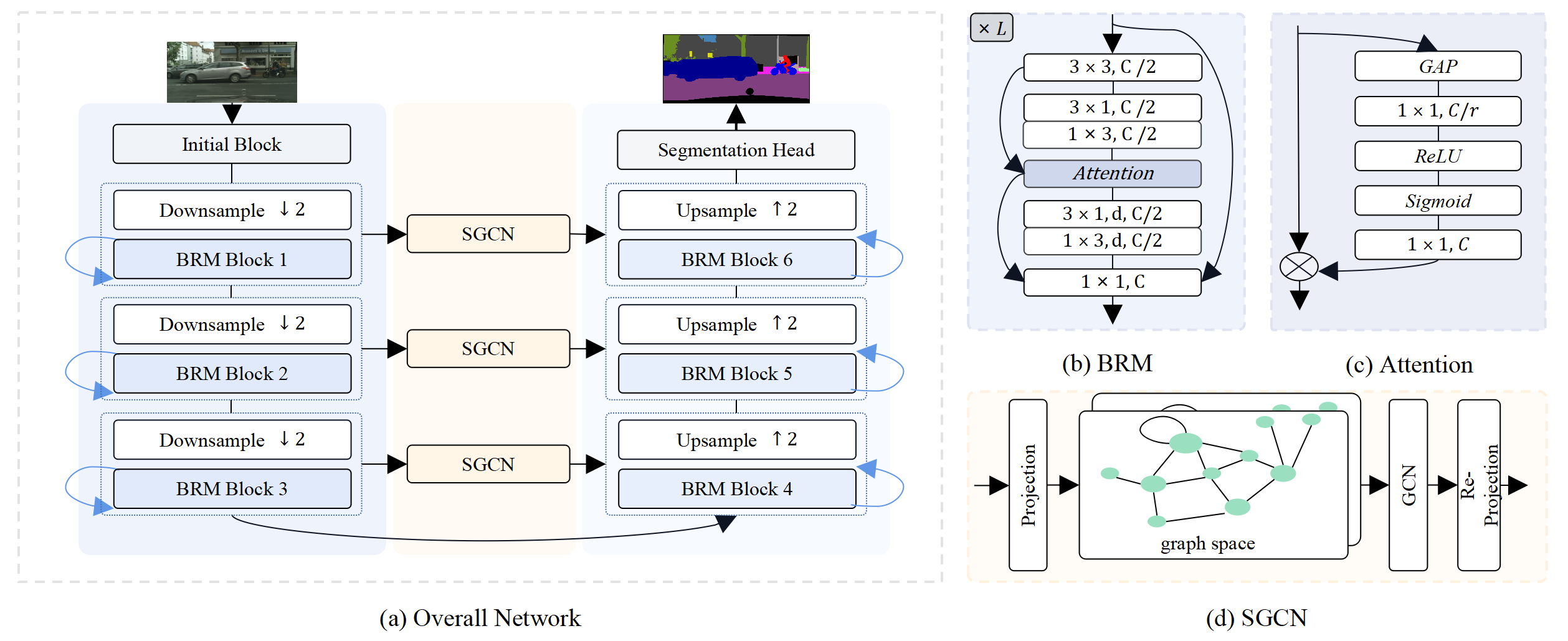}}
	\caption{The detailed model architecture of our proposed 
 MFPNet, which is composed of 
 a Bottleneck Residual Module (BRM), a Simple Graph Convolutional Network (SGCN), and an Atrous Spatial Pyramid Pooling module (ASPP).}
	\label{Figure 2}
\end{figure*}


\section{Related Work}
\label{sec: Related Work}

\subsection{Semantic Segmentation}
\label{ssec:sub2.1}

Contemporary approaches to semantic segmentation can be broadly categorized into two groups: large-scale models and lightweight models.

The noteworthy large-scale models include CNN-based models like DANet~\cite{fu2019dual}, integrating 
Non-local attention, Transformer-based models such as 
SegFormer~\cite{xie2021segformer}, combining the local vision of CNNs with the global perception of Transformers, and Diffusion-based models such as 
DDPM~\cite{lai2023denoising}, generating predictions by employing iterative noise addition and the subsequent 
denoising.

Representative lightweight models include models such as BiseNet~\cite{yu2021bisenet} and DFANet~\cite{li2019dfanet}, built upon lightweight backbones. 
Furthermore, real-time strategies such as ERFNet~\cite{romera2017erfnet} and FBSNet~\cite{gao2022fbsnet} involve transplanting and optimizing effective modules from large models to construct entire models. 
However, it is worth noting that these methods often 
encounter a marginal performance gap. 
As a result, 
some researchers are dedicated 
to exploring methods 
to optimize the potential of these constrained features. 
Strategies include deploying dilated convolutions to expand the receptive fields, augmenting network width, and incorporating attentions~\cite{hu2018squeeze}.

\subsection{Graph Convolutional Networks}
\label{ssec:sub2.2}

Graph Convolutional Networks~\cite{kipf2016semi} employ a layer-wise \textit{message-passing} function achieved 
through the fusion of linear transformations. 

This property enables them to effectively model long-range dependencies while retaining local details, beneficial for resolution recovery.

The latent embedding features for the ${l}$-th layer $H^{\left( l \right)}$, responsible for aggregating information within one-hop neighborhoods, can be formally defined as:
\begin{equation}
    {H^{\left( l \right)}} = \sigma \left( {\Hat {A}  {H^{\left( {l - 1} \right)}}{\Theta ^{\left( {l - 1} \right)}}} \right),
\end{equation}
where $\sigma$ denotes an activation function, ${\Theta ^{\left( {l - 1} \right)}} \in {\mathbb{R}^{d \times d}}$  denotes the trainable weight, $d$ represents the dimension of graph space, and $\Hat {A}$ signifies 
the normalization of the adjacency matrix $A\in {\mathbb{R}^{n \times n}}$, $n$ denotes the number of nodes, including self-constructing loops~\cite{liu2020self}:
\begin{equation}
    \Hat {A}   = {D^{ - \frac{1}{2}}}\left( {A + I} \right){D^{\frac{1}{2}}},
\end{equation}
\begin{equation}
    {D_{ii}} = {\sum\nolimits_j^{} {\left( {A + I} \right)} _{ij}}.
\end{equation}
where $D$ means the degree matrix and $I$ is the identity matrix.

\section{Methodology}
\label{sec:Methodology}

\vspace{-0.5em}
\subsection{Overall Architecture}
\label{ssec:sub3.1}

As illustrated 
in Fig.~\ref{Figure 2}a, our proposed model incorporates a comprehensive Encoder-Decoder framework. The gradual symmetrical stage in the feature fusion aids in supplementing the details lost during encoding. 
Notably, simple GCNs are used in the fusion process to model the feature information around adjacent elements in the latent space. 

Specifically, given an input image $X \in {\mathbb{R}^{3 \times H \times W}}$, the first step is to pass the input image through the Initial Block, which is composed of three consecutive $3 \times 3$ convolution layers. 
These convolution layers are utilized to comprehensively extract the initial feature $F \in {\mathbb{R}^{C \times \mathop H\limits^ \sim   \times \mathop W\limits^ \sim  }}$. 
Then, the Encoder consists of 
three downsampling layers of varying scales, each followed by 
a Bottleneck Residual Module (BRM) block, denoted as $Block_{1,2,3}$ generating ${E_i} \in {\mathbb{R}^{{C_i} \times {H_i} \times {W_i}}}$, $\{ i = 1,2,3\}$, respectively. 

The Decoder mirrors 
the Encoder, maintaining the same scale alignment, with the BRM Blocks being sequentially labeled 
as $Block_{4,5,6}$, generating ${D_j} \in {\mathbb{R}^{{C_j} \times {H_j} \times {W_j}}}$, $\{ j = 1,2,3\}$. 
Each BRM Block contains a set of ${L_n}$ BRMs, $\{ n = 1,...,6\}$. 
For efficient feature reuse and enhancing 
the network, each block employs a residual connection. 

The middle information propagator, composed of series of multi-scale Simple GCN, takes 
the input from the corresponding scale of the Encoder,  
which is then transformed into the graph space ${G_i} \in {\mathbb{R}^{{D_i} \times {N_i}}}$, $\{ i = 1,2,3\}$ through projection. 

As shown in Fig.~\ref{Figure 2}d, after GCN processing, the long-range feature information ${G_i}$ is integrated and then re-projected back to the conventional feature space to obtain ${M_i} \in {\mathbb{R}^{{C_i} \times {H_i} \times {W_i}}}$, $\{ i = 1,2,3\}$. Finally, the Segmentation Head is used for the ultimate optimization of the output features before pixel classification.

\begin{table}[htbp]
\caption{Details of the experiment settings.} 
\begin{center}
\setlength\tabcolsep{8 pt}
\renewcommand{\arraystretch}{1.2}
\scalebox{0.9}{
\begin{tabular}{l|c|c}
\toprule[0.35mm]
\hline
\rowcolor{gray!20}Dataset & \multicolumn{1}{c|}{Cityscapes~\cite{cordts2016cityscapes}}  & CamVid~\cite{brostow2008segmentation} \\ \hline
\multirow{2}{*}{Learning Rate} & \multicolumn{2}{c}{$Poly$ ( ${lr = l{r_{in}} \times {\left( {1 - \frac{{iter}}{{{total_{iter}}}}} \right)^{0.9}}}$)}                 \\ \cline{2-3} 
                               & \multicolumn{1}{c|}{$l{r_{in}} = 4.5 \times {10^{ - 2}}$}      & $l{r_{in}} = 1 \times {10^{ - 3}}$       \\ \hline
\multirow{2}{*}{Optimizer}     & \multicolumn{1}{c|}{$SGD$}         &  $Adam$      \\ \cline{2-3} 
                               & \multicolumn{1}{c|}{$ wd = 1 \times {10^{ - 5}}$} &  $ wd = 2 \times {10^{ - 5}}$      \\ \hline
Loss Function                  & \multicolumn{2}{c}{$CrossEntry$ $Loss$ $Function$}                    \\ \hline
Resolution            & \multicolumn{1}{c|}{$ 512 \times 1024 $}            & $ 360 \times 480 $           \\ \hline

\bottomrule[0.35mm]
\end{tabular}
}
\label{Table 1}
\end{center}
\vspace{-2.0em}
\end{table}

\begin{table*}[htbp]
\caption{Quantitative comparison results with the state-of-the-art methods on the Cityscapes~\cite{cordts2016cityscapes} and CamVid~\cite{brostow2008segmentation}. $Para.$ represents $Parameters$, ``$*$" means more than a single GPU, and ``$-$" denotes not provided in the original reference.}
\begin{center}
\setlength\tabcolsep{4 pt}
\renewcommand{\arraystretch}{1.1}
\scalebox{0.8}{
\begin{tabular}{l|ccc|ccccccc}
\toprule[0.35mm]
\hline
\rowcolor{gray!20}\multirow{2}{*}{Methods} & \multirow{2}{*}{Backbone} & \multirow{2}{*}{Para.(M) ${\downarrow}$} & \multirow{2}{*}{GPU} & \multicolumn{3}{c}{Cityscapes~\cite{cordts2016cityscapes}} &  
& \multicolumn{3}{c}{CamVid~\cite{brostow2008segmentation}} \\ \cline{5-7} \cline{9-11}
&  &  &  & \multicolumn{1}{c}{FLOPs (G) ${\downarrow}$} & \multicolumn{1}{c}{Speed (FPS) ${\uparrow}$} & mIoU (\%) ${\uparrow}$ &  
& \multicolumn{1}{c}{Resolution} & \multicolumn{1}{c}{Speed (FPS) ${\uparrow}$} & mIoU (\%) ${\uparrow}$ \\ \hline
DeepLab-V3+ ~\cite{chen2018encoder}
& MobileNet-V2 & 62.70 & $*$ & \multicolumn{1}{c}{2032.3} & \multicolumn{1}{c}{1.2}  & 80.9     &  & \multicolumn{1}{c}{-}    & \multicolumn{1}{c}{-}     & -     \\ 

SegFormer~\cite{xie2021segformer} 
& MiT-B5  & 84.70  & $*$  & \multicolumn{1}{c}{1447.6} & \multicolumn{1}{c}{2.5}  & 84.0     &  
& \multicolumn{1}{c}{-}           & \multicolumn{1}{c}{-}      &  -  \\ 
                         
DDPS-SF~\cite{lai2023denoising} & MiT-B5  & 122.80  
& $*$  & \multicolumn{1}{c}{-}      & \multicolumn{1}{c}{-}    &82.4      &  & \multicolumn{1}{c}{-}           & \multicolumn{1}{c}{-}   &-  \\  \hline
                         
ENet~\cite{paszke2016enet}& No  & 0.36  & Titan X  
& \multicolumn{1}{c}{3.8}      & \multicolumn{1}{c}{135}      & 58.3     &  & \multicolumn{1}{c}{$360 \times 480$}   & \multicolumn{1}{c}{62} & 51.3 \\ 
                         
ERFNet~\cite{romera2017erfnet}& No   & 2.10  & Titan X  & \multicolumn{1}{c}{25.8} & \multicolumn{1}{c}{42}   & 68.0     &  & \multicolumn{1}{c}{$720 \times 960$}   & \multicolumn{1}{c}{139}  &67.7  \\ 
                         
ICNet~\cite{zhao2018icnet} & PSPNet-50   & 26.50   & Titan X & \multicolumn{1}{c}{28.3}      & \multicolumn{1}{c}{30}      & 69.5     &  
& \multicolumn{1}{c}{$720 \times 960$} & \multicolumn{1}{c}{34}  
& 67.1 \\ 

DABNet~\cite{li2019dabnet} & No   & 0.76   & 1080Ti & \multicolumn{1}{c}{10.5}      & \multicolumn{1}{c}{104}      & 70.1     &  
& \multicolumn{1}{c}{$360 \times 480$}    & \multicolumn{1}{c}{146}  & 66.4\\ 
                         
FBSNet~\cite{gao2022fbsnet} & No   & 0.62   & 2080Ti   & \multicolumn{1}{c}{9.7}      & \multicolumn{1}{c}{90}    & 70.9     &  
& \multicolumn{1}{c}{$360 \times 480$}     & \multicolumn{1}{c}{120} & 68.9 \\ 
                         
SGCPNet~\cite{hao2022real} & MobileNet  & 0.61   & 1080Ti  
& \multicolumn{1}{c}{4.5}      & \multicolumn{1}{c}{103}      & 70.9     &  & \multicolumn{1}{c}{$720 \times 960$}  & \multicolumn{1}{c}{278} & 69.0  \\ 
                         
DFANet~\cite{li2019dfanet}& Xception   & 7.80   & Titan X  & \multicolumn{1}{c}{3.4}      & \multicolumn{1}{c}{100}      & 71.3     &  & \multicolumn{1}{c}{$720 \times 960$}           & \multicolumn{1}{c}{120}  & 64.7  \\
                        
DSANet~\cite{elhassan2021dsanet}& No   & 3.47  & 1080Ti   & \multicolumn{1}{c}{37.4}      & \multicolumn{1}{c}{34}      & 71.4     &  & \multicolumn{1}{c}{$360\times480$} & \multicolumn{1}{c}{75}  & 69.9 \\ \hline

\multirow{2}{*}{\textbf{MFPNet}}        & \multirow{2}{*}{No}         & \multirow{2}{*}{1.00}      & \multirow{2}{*}{V100}    & \multicolumn{1}{c}{\multirow{2}{*}{18.6}} & \multicolumn{1}{c}{\multirow{2}{*}{106}} & \multirow{2}{*}{\textbf{71.5}} & \multirow{2}{*}{} 
& \multicolumn{1}{c}{$360\times480$}  & \multicolumn{1}{c}{163}  & 68.1     \\ 
&        &      &     & \multicolumn{1}{c}{}    & \multicolumn{1}{c}{} &                   &                   & \multicolumn{1}{c}{$720\times960$}   & \multicolumn{1}{c}{98}   & 69.2  \\ \hline

\bottomrule[0.35mm]
\end{tabular}
}
\label{Table 2}
\end{center}
\vspace{-1.5em}
\end{table*}

\subsection{Bottleneck Residual Module (BRM)}
\label{ssec:sub3.2}

As shown in Fig.~\ref{Figure 2}b, the entirety of BRM is a residual structure which makes it possible to deepen the network. 
The head employs a bottleneck $3 \times 3$ convolutional layer for minimizing computational load and parameter burden. 
Then, the intermediate layer comprises of two sets of factorized convolutions, with the second set incorporating dilated convolutions to amplify the receptive fields. 
Notably, an attention mechanism~\cite{hu2018squeeze} is interposed between them, as depicted in Fig.~\ref{Figure 2}c, which serves to intensify feature expression. Finally, a $1 \times 1$ convolution is employed to recover the channels.

\subsection{Simple Graph Convolutional Network (SGCN)}
\label{ssec:sub3.3}

As shown in Fig.~\ref{Figure 2}d, the features ${E_i} \in {\mathbb{R}^{{C_i} \times {H_i} \times {W_i}}}$, $\{ i = 1,2,3\}$ received from the Encoder are first mapped into the graph space ${\Omega _g}$: 
\begin{equation}
    {G_i} = (A_i, X_i) = \Phi \left( {{E_i}} \right) \in {\mathbb{R}^{{d_i} \times {n_i}}},
\end{equation}
where ${n_i} = {h_i} \times {w_i} \le {H_i} \times {W_i}$ denotes the number of nodes, $\Phi$ is the projection operation, and $d_i$ represents the dimension of graph space. ${X_i}$ are the node features, and the adjacency matrix ${A_i} = f\left( {\delta \left( {{G_i}} \right),\psi {{\left( {{G_i}} \right)}^T}} \right)$ is harnessed to diffuse and propagate feature information among nodes. 

For each node, the mapping comprises two components: the $n$ nearest neighbor pixels ($x_i$, $y_i$), $i < N$, and the associated weight $q_i$. 
It entails updating subsequent vertex features based on edge attributes. The outcome of graph convolution is represented as: 
\begin{equation}
    {M_i} = \Hat {A}  \left( {g\left( {{G_i}} \right)} \right){Q_i},
\end{equation}
where $Q_i$ is the learnable weight matrix, and $g$ refers to 
graph convolution. 

Note that, similar feature propagation ideas have been proposed in~\cite{ding2019boundary,liu2022instance}. 
However, the boundary-aware requirement in object identification often demands high resolution. Instead, our method does not require it and excels with a rough guide. This differentiates our approach from the existing ones.

\begin{figure}
     \centering
     \begin{subfigure}[b]{0.45\textwidth}
         \centering
         \includegraphics[width=\textwidth]{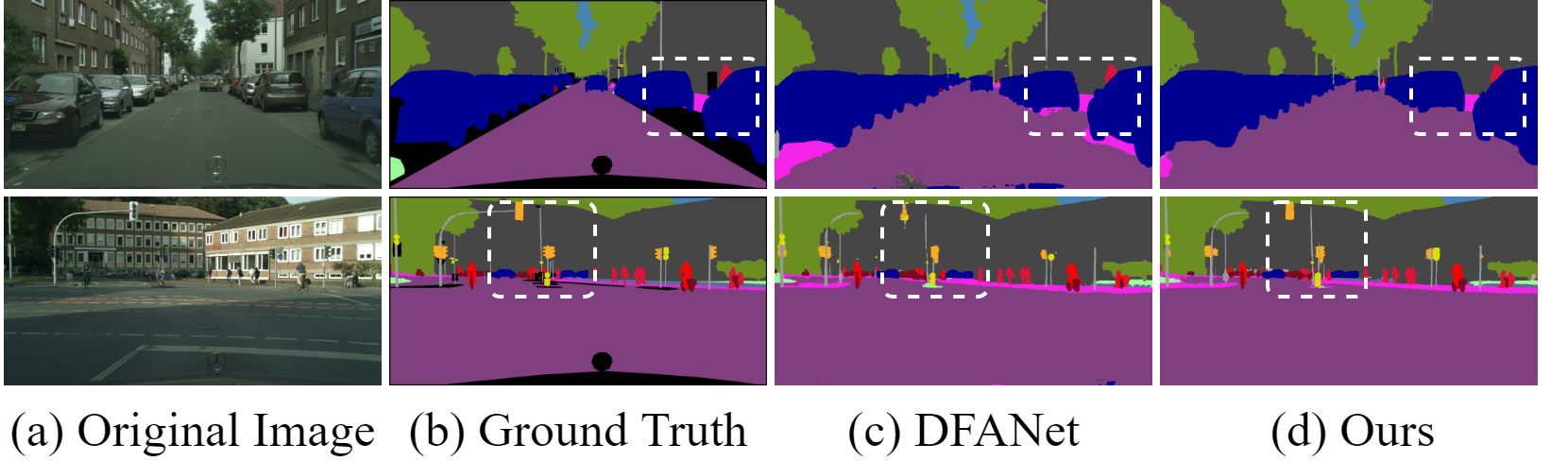}
         \caption{Results on Cityscapes~\cite{cordts2016cityscapes} (compared with DFANet~\cite{li2019dfanet}) }
         \label{Figure 3}
     \end{subfigure}
      \vfill
     \begin{subfigure}[b]{0.45\textwidth}
         \centering
         \includegraphics[width=\textwidth]{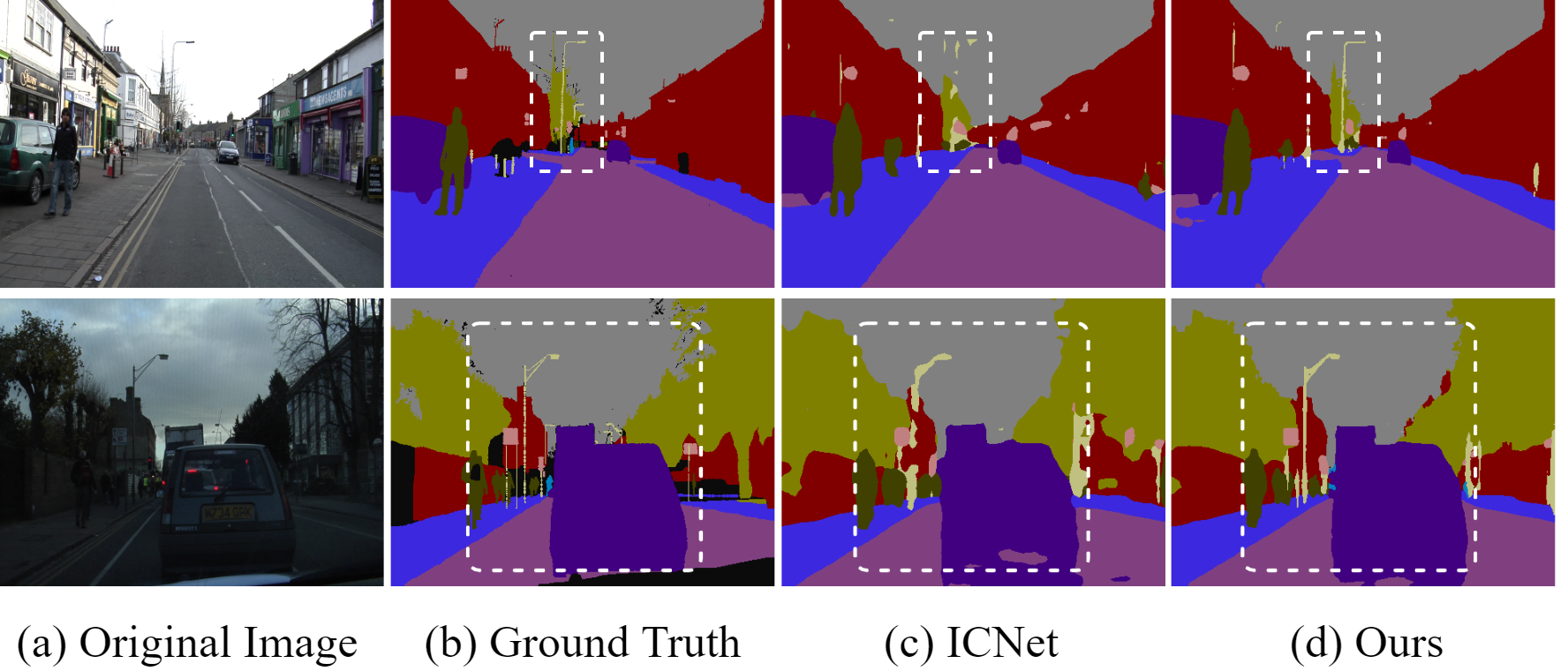}
         \caption{Results on CamVid~\cite{brostow2008segmentation} (compared with ICNet~\cite{zhao2018icnet})}
         \label{Figure 4}
     \end{subfigure}
     \caption {Visual qualitative results obtained on Cityscapes~\cite{cordts2016cityscapes} and CamVid~\cite{brostow2008segmentation}. Note particularly the areas highlighted with the white dotted boxes.}
\end{figure}

\section{Experiments}
\label{sec:experiments}

\subsection{Implementation Details}
\label{ssec:sub4.1}

We implement our approach using PyTorch 1.8.1 with Python 3.6 and Cuda 10.1. 
We validate our MFPNet on two public semantic segmentation datasets, Cityscapes~\cite{cordts2016cityscapes} and CamVid~\cite{brostow2008segmentation}, of
which the training details are slightly different, as detailed in Table~\ref{Table 1}. 
We compare our MFPNet with the SOTA segmentation models in terms of
Params, FLOPs, FPS, and mIoU under the different image resolutions
listed in the tables. 
Experimental results of Cityscapes and CamVid are shown in Tables~\ref{Table 2} and~\ref{Table 3}, respectively. 
All experiments were conducted with one Tesla V100 GPU card. 


\subsection{Results}
\label{ssec:sub4.2}

As shown in Table~\ref{Table 2}, large models achieve remarkable accuracy but come with substantial demands on hardware resources and exhibit limited inference speed. 
Among the lightweight models, DSANet~\cite{elhassan2021dsanet} has achieved an accuracy close to ours 
at the expense of a threefold increase in parameters, and an additional 20G FLOPs. 
Other methods exhibit slightly lower accuracy than ours. 
The data substantiates that our approach achieves a more favorable trade-off between accuracy, size, and speed. 
The outcomes of visualization on the Cityscapes dataset~\cite{cordts2016cityscapes} are illustrated in Fig.~\ref{Figure 3}.

The CamVid dataset~\cite{brostow2008segmentation}, as presented in Table~\ref{Table 2}, boasts a smaller resolution and fewer samples, effectively underscoring the generalizability of our model. The visualized outcomes are depicted in Fig.~\ref{Figure 4}.

It can be seen 
that, with our proposed MFPNet, the boundary details of objects are well recovered thanks to the feature propagation strategy.
Additionally, the long-range dependencies of multi-scale GCN have contributed to a reduction in misclassification of small objects such as traffic lights.

\subsection{Ablation Studies}
\label{ssec:sub4.3}


\textbf{Ablation Study on SGCN.} 
As shown 
in Table~\ref{Table 3}, the inclusion of SGCN modeling long-range pixel dependencies, reduces misclassification and yields enhanced model performance on both datasets. 
The improvement is notable, ascending from 69.9\% mIoU to 71.5\% mIoU on Cityscapes and from 66.2\% mIoU to 68.1\% mIoU on CamVid. 
Remarkably, these gains come at a modest cost, with parameters merely increasing by 200k and FLOPs experiencing a rise of less than 1G.

\begin{table}[htbp]
\caption{Ablation studies on the Simple Graph Convolutional Network (SGCN) on the datasets Cityscapes and CamVid. $P.$, $F.$, and $val$ denote $Parameters$, $FLOPs$, and $validation$.} 
\begin{center}
\setlength\tabcolsep{3 pt}
\renewcommand{\arraystretch}{1.2}
\scalebox{0.8}{
\begin{tabular}{l|ccccccccccc}
\toprule[0.35mm]
\hline
\rowcolor{gray!20}\multirow{6}{*}{\rotatebox{90}{MFPNet}} & \multirow{3}{*}{SGCN} & \multirow{3}{*}{P. (K)} & \multicolumn{3}{c}{Cityscapes~\cite{cordts2016cityscapes}}            &  & \multicolumn{3}{c}{CamVid~\cite{brostow2008segmentation}}                                                     \\ \cline{4-6} \cline{8-10}
                       &                       &                     & \multicolumn{1}{c}{\multirow{2}{*}{F. (G)}} & \multicolumn{2}{c}{mIoU (\%)}       &  & \multicolumn{1}{c}{\multirow{2}{*}{F. (G)}} & \multicolumn{2}{c}{mIoU (\%)}       \\ \cline{5-6} \cline{9-10} 
                       &                       &                     & \multicolumn{1}{c}{}                      & \multicolumn{1}{c}{val} & test &  & \multicolumn{1}{c}{}                      & \multicolumn{1}{c}{val} & test \\ \cline{2-10} 
                       & \XSolidBrush      & 843.20                    & \multicolumn{1}{c}{17.8}                    & \multicolumn{1}{c}{71.2}  & 69.9  &  
                       & \multicolumn{1}{c}{5.8}                      & \multicolumn{1}{c}{88.3}    & 66.2     \\ \cline{2-10} 
                       & \Checkmark        & 1001.57                    & \multicolumn{1}{c}{18.6}                     & \multicolumn{1}{c}{73.0}  & 71.5  &  
                       & \multicolumn{1}{c}{6.2}                      & \multicolumn{1}{c}{91.7}    & 68.1        \\ \hline

\bottomrule[0.35mm]
\end{tabular}
}
\label{Table 3}
\end{center}
\end{table}


\textbf{Ablation on Segmentation Head.} 
Our experiment encompasses various segmentation heads, \textit{i.e.}, PSP, SE, and ASPP, as shown in Table~\ref{Table 4}. 
As for ASPP, we configure the reduction as 4 to optimize model size. 
All these alternatives have contributed to accuracy enhancements. Remarkably, ASPP exhibits the most promising outcome, achieving 
lies in parallel multi-rate dilated convolutions, which preserve the image resolution and better perceive objects of different size.

\begin{table}[htbp]
\caption{Ablation studies on the Segmentation Head on the CamVid ($360 \times 480$).}
\begin{center}
\setlength\tabcolsep{7 pt}
\renewcommand{\arraystretch}{1.1}
\scalebox{0.8}{
\begin{tabular}{l|l|ccccccccc}
\toprule[0.35mm]
\hline
\rowcolor{gray!20} \multicolumn{2}{c|}{Model} & Para. (K) & FLOPs (G) & mIoU (\%)\\ \hline
\multicolumn{1}{c}{\multirow{4}{*}{\rotatebox{90}{MFPNet}}} 
& -  & 999.70      & 5.80      & 66.1     \\ 

\multicolumn{1}{c}{}        
& +SE   & 999.85   & 5.82      & 66.8 ( +0.7 )    \\ 

\multicolumn{1}{c}{}  
& +PSP  & 999.92   & 5.82      & 67.5 ( +1.4 )     \\ 

\multicolumn{1}{c}{}         
& +ASPP & 1001.57      & 6.20      & 68.1 ( +2.0 )    \\ 

\hline
\bottomrule[0.35mm]
\end{tabular}
}
\label{Table 4}
\end{center}
\vspace{-2.0em}
\end{table}

\textbf{Ablation on the number of BRMs and Dilated Rate.} 
As illustrated in Table~\ref{Table 5}, the trend emerges – higher module quantities correspond to improved segmentation results. This is attributed to the fact that heightened network depth correlates with the more advanced semantic context. Moreover, elevating the dilated rates widens the receptive fields engendering more comprehensive features. 
In comparison to one module per Block, the final version integrating three BRMs per Block leads to a remarkable 4.2\% enhancement.

\begin{table}[htbp]
\caption{Ablation studies on the number of BRMs and Dilated Rate on the CamVid ($360 \times 480$). $L_n$ corresponds to the Block number as introduced in Section~\ref{ssec:sub3.1}.}
\begin{center}
\setlength\tabcolsep{7 pt}
\renewcommand{\arraystretch}{1.15}
\scalebox{0.8}{
\begin{tabular}{l|ccccc}
\toprule[0.35mm]
\hline
\rowcolor{gray!20}& $L_1 = L_6$  & $L_2 = L_5$  & $L_3 = L_4$      & mIoU (\%)  ${\uparrow}$ \\ \hline
\multirow{3}{*}{\rotatebox{90}{MFPNet}} 
& (1)    & (1)    &  (1)                      & 63.9     \\ 
& (2, 2, 2)    & (2, 2, 2)    &  (2, 2, 2)    & 66.8 ( +2.9 )    \\  
& (2, 2, 2)    & (4, 8, 16)    &  (4, 8, 16)    & 68.1 ( +4.2 )    \\ \hline
\bottomrule[0.35mm]
\end{tabular}
}
\label{Table 5}
\end{center}
\end{table}

\vspace{-2.5em}
\section{Conclusion}
\label{sec:conclusionn}

\vspace{-0.5em}
This paper proposes a light yet efficient network MFPNet for real-time lightweight semantic segmentation. 
Intermediate information undergoes propagation via embedding the multi-scale simple GCNs, to model latent dependencies between long-range objects, a critical advantage for segmentation accuracy. 
The incorporation of meticulously designed bottleneck residual modules serve to deepen the network for powerful semantic insight. 
Finally, the segmentation head employs the classical ASPP, requiring minimal effort while achieving 
prediction enhancements.

\vspace{-0.5em}

\bibliographystyle{IEEEbib}
\bibliography{strings,refs}

\end{document}